\documentclass[letterpaper, 10 pt, conference]{ieeeconf}  

\IEEEoverridecommandlockouts                              

\usepackage{cite}
\usepackage{amsmath,amssymb,amsfonts}
\usepackage{algorithm}
\usepackage{algpseudocode}
\usepackage{graphicx}
\usepackage{textcomp}
\usepackage{xcolor}
\usepackage{algorithm}
\usepackage{threeparttable}
\usepackage{algpseudocode}
\usepackage[table,xcdraw]{xcolor}
\usepackage{float}
\usepackage{booktabs}
\usepackage{hyperref}
\usepackage{afterpage}
\usepackage{listings}
\usepackage{balance}
\usepackage{wrapfig}
\usepackage{caption}
\usepackage{multirow}
\usepackage{subcaption}
\title{\LARGE \bf
Real-Time Synchronized Interaction Framework for Emotion-Aware Humanoid Robots*
}
\author{Yanrong Chen$^{1}$ and Xihan Bian$^{2}$
\thanks{*This work was supported by Xi’an Jiaotong-Liverpool University under internal research funding. The NAO humanoid robot was provided by the university. The large language model resources were supported by Alibaba Cloud. Project guidance was provided by Dr. Xihan Bian. Project Website: \url{https://cyr1213.github.io/ReSIn-HR/}}
\thanks{$^{1}$Yanrong Chen is with the Department of Computing, Xi’an Jiaotong-Liverpool University, Suzhou, China.
        {\tt\small chenyanrong2025@163.com}}%
\thanks{$^{2}$Xihan Bian is with the Department of Computing, Xi’an Jiaotong-Liverpool University, Suzhou, China.
        {\tt\small xihan.bian@xjtlu.edu.cn}}%
}

\begin{document}

\maketitle
\thispagestyle{empty}
\pagestyle{empty}

\begin{abstract}
    As humanoid robots increasingly introduced into social scene, achieving emotionally synchronized multimodal interaction remains a significant challenges. To facilitate the further adoption and integration of humanoid robots into service roles, we present a real-time framework for NAO robots that synchronizes speech prosody with full-body gestures through three key innovations: (1) A dual-channel emotion engine where large language model (LLM) simultaneously generates context-aware text responses and biomechanically feasible motion descriptors, constrained by a structured joint movement library; (2) Duration-aware dynamic time warping for precise temporal alignment of speech output and kinematic motion keyframes; (3) Closed-loop feasibility verification ensuring gestures adhere to NAO’s physical joint limits through real-time adaptation. Evaluations show 21\% higher emotional alignment compared to rule-based systems, achieved by coordinating vocal pitch (arousal-driven) with upper-limb kinematics while maintaining lower-body stability. By enabling seamless sensorimotor coordination, this framework advances the deployment of context-aware social robots in dynamic applications such as personalized healthcare, interactive education, and responsive customer service platforms.
\end{abstract}


\section{INTRODUCTION}

As robotic systems increasingly permeate social domains including healthcare, education, and service industries, the demand for emotionally resonant and temporally coordinated human-robot interaction (HRI) has become critical. Studies show that gesture-speech synchronization can significantly enhance users’ perception of empathy and engagement \cite{breazeal2003toward}. Despite this, achieving real-time co-speech gesture generation in physically embodied systems remains an open challenge, constrained by three core factors: (1) the high-dimensional complexity of motion dynamics \cite{zhang2024kinmo}, (2) variability in speech prosody and affective shifts \cite{wu2019dual}, and (3) strict biomechanical constraints of robotic platforms such as NAO \cite{nao2018whitepaper}.

Early gesture synthesis systems primarily relied on rule-based mappings between predefined linguistic cues and gesture templates\cite{bhattacharya2021speech2affectivegestures}, which were limited in adaptability. Data-driven approaches later introduced statistical learning from curated datasets\cite{kucherenko2019analyzing}, enabling more flexible outputs but often remained constrained to 2D gestures or static mappings. The advent of deep generative models, such as Human Motion Diffusion Models (HMDMs) \cite{ho2020denoising, tevet2022human}, allowed for temporally coherent and stylistically consistent full-body motion synthesis. Techniques like DiffSHEG \cite{chen2024diffsheg} and MoFusion \cite{tevet2022human} leveraged affect-conditioned diffusion or transformer-based pipelines to produce expressive gesture sequences, yet they often rely on pre-segmented emotion labels and are not optimized for real-time interaction.

Recent developments in using large language models (LLMs) for embodied control \cite{liang2023code,brohan2023rt} have demonstrated that linguistic input can be mapped directly to low-level control actions. Building on this foundation, works such as MotionGPT \cite{zhang2024motiongpt} explore LLMs for high-level motion synthesis. However, these frameworks typically focus on semantic alignment and omit real-time emotional adaptation or physical constraint integration.
\begin{figure}[t]
\centering
\includegraphics[width=1\linewidth]{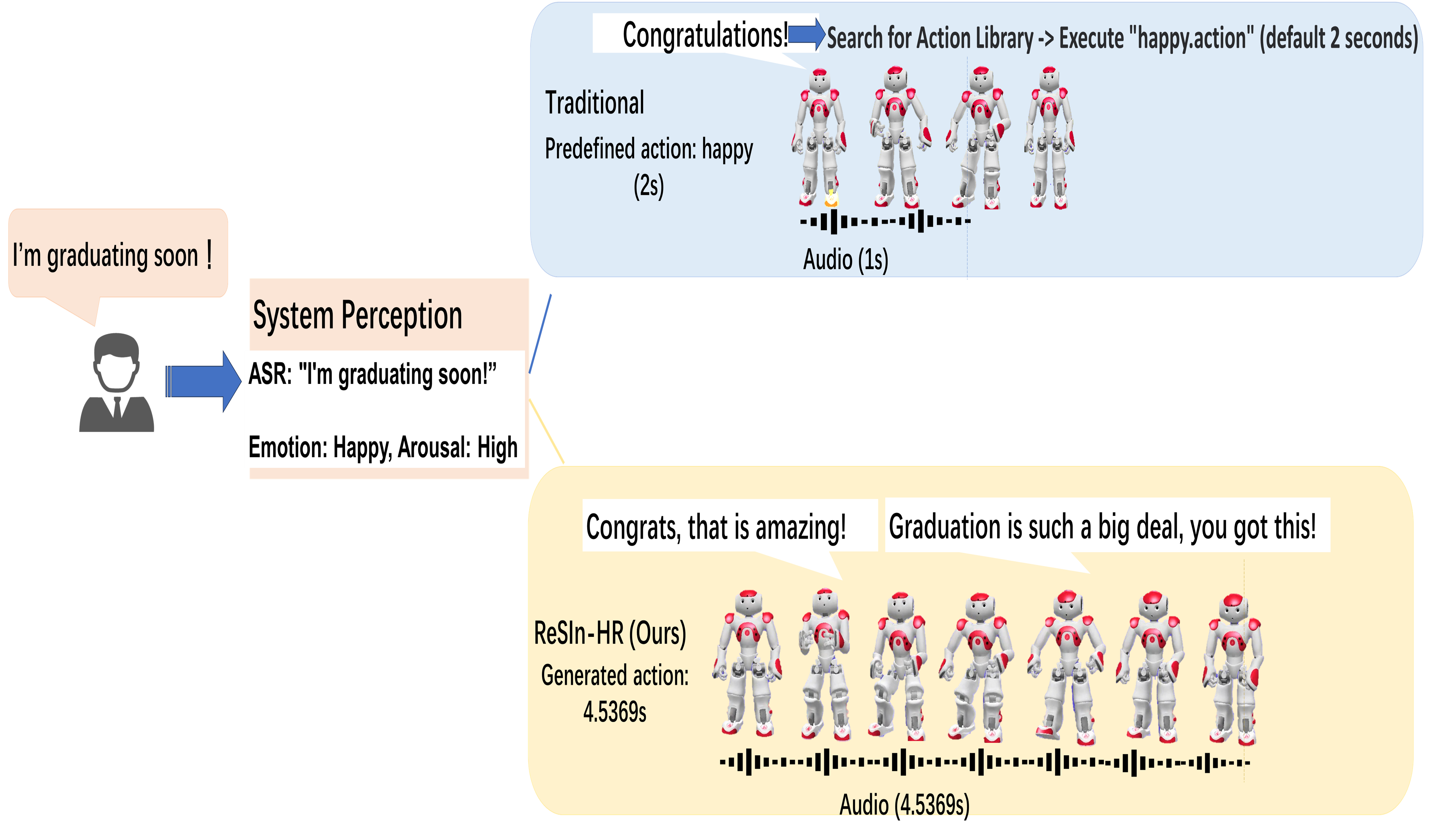} 
\caption{Traditional gesture systems play static actions disconnected from speech timing. ReSIn-HR generates dynamic, emotionally aligned gestures that synchronize with both content and prosody.}
\label{fig:gesture_sync_comparison}
\vspace{-0.5cm}
\end{figure}
To address these challenges, this work introduces \textbf{ReSIn-HR} (\textbf{Re}al-time \textbf{S}ynchronized \textbf{In}teraction for \textbf{H}umanoid \textbf{R}obots), a novel framework for synchronized speech-gesture generation. Our approach integrates real-time speech emotion recognition (SER) with adaptive gesture planning, offering the following three core innovations:
\begin{figure*}[t]
\centering
\includegraphics[width=\linewidth]{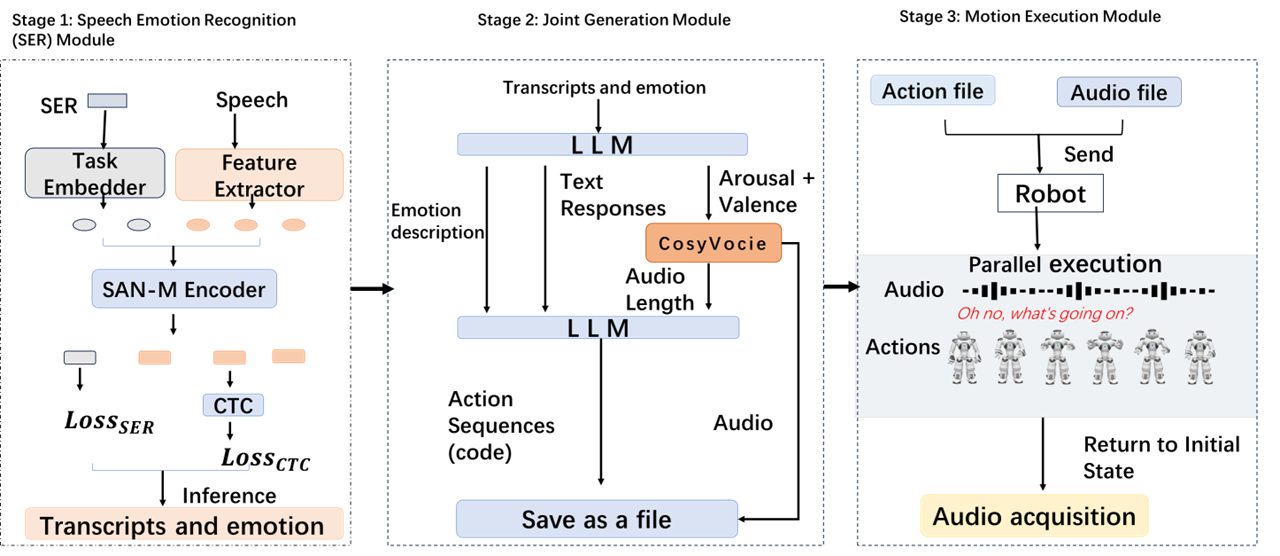}
\caption{System architecture of ReSIn-HR for multimodal interaction.}
\label{fig:overview}
\vspace{-0.5cm}
\end{figure*}
\begin{enumerate}
    \item \textbf{Dual-Stream Emotion Processing} \newline
    A pipeline that extracts both linguistic and prosodic cues to inform gesture selection, enabling emotion-driven motion adaptation.

    \item \textbf{Duration-Aware Synchronization} \newline
    A mechanism that dynamically predicts speech duration and adjusts gesture keyframes to minimize timing misalignment.

    \item \textbf{Biomechanical Constraint Verification} \newline
    A real-time module that ensures generated gestures adhere to the robot’s physical limitations, preventing unnatural movements or execution failures.
\end{enumerate}

\noindent
\textbf{Experimental validation} with the NAO robot demonstrates that the ReSIn-HR framework achieves a \textbf{21\% higher emotional alignment} compared to a rule-based NAO system. By unifying affective signal processing, temporal coordination, and hardware-aware motion synthesis, ReSIn-HR advances the state of the art in socially adaptive human-robot interaction (HRI), enabling robots to produce contextually appropriate gestures that align with both linguistic content and emotional tone in real time.

Fig~\ref{fig:gesture_sync_comparison} compares a rule-based system with our ReSIn-HR framework. While the traditional method plays a fixed “happy” gesture disconnected from speech dynamics, our system generates emotionally appropriate and temporally synchronized gestures that evolve naturally with the utterance. This demonstrates ReSIn-HR’s ability to deliver expressive, speech-aware motion, enhancing the fluidity and empathy of robot interaction.

\section{RELATED WORK}
\textbf{Emotion-Aware Gesture Generation:}  
Early approaches to gesture synthesis relied on rule-based mappings~\cite{marsella2013virtual,poggi2005greta} and statistical models~\cite{cassell1994animated,habibie2021learning}, which suffered from limited generalizability due to handcrafted rules and small datasets. With the advent of 3D human pose estimation~\cite{loper2023smpl,zhang2023pymaf} and deep temporal models~\cite{pavlakos2019expressive,boukhayma20193d}, data-driven methods emerged, learning gesture patterns from audio, text, and speaker identity. Emotion-conditioned models introduced valence-arousal modulation and affective GANs~\cite{qi2024weakly}, but typically rely on scripted datasets and struggle in real-time, spontaneous interaction. In addition, gesture timing has been explored via prosodic cues~\cite{kucherenko2020genea} and latent alignment~\cite{li2021audio2gestures}, though robustness remains an issue under hardware latency and spontaneous speech. Our work builds upon these foundations by integrating real-time emotion detection with gesture generation, using speech duration and emotion shifts as soft anchors for synchronization.

\textbf{LLMs for Motion and Gesture Synthesis:}  
Recent progress in robot control has leveraged large language models (LLMs) to bridge language and action. Code-as-Policies~\cite{liang2023code} and RT-2~\cite{brohan2023rt} show that LLMs can translate textual prompts into executable policies. MotionGPT~\cite{zhang2024motiongpt} and MoFusion~\cite{tevet2022human} extend this to text-to-motion generation by embedding prompts into continuous motion spaces. However, most models focus on semantic alignment, with limited emotional modulation. Emotion-aware gesture generation~\cite{bhattacharya2021speech2affectivegestures,chen2024diffsheg} often depends on fixed emotion categories and external classifiers, while motion feasibility is treated as a post-processing step. Prompt engineering for robotic planning has introduced symbolic and programmatic structures, but lacks integration of dynamic features like affect curves or joint constraints. We extend this line by encoding multimodal constraints—including valence-arousal over time, semantic intent, and joint limits—directly into LLM prompts, enabling expressive and physically valid gesture generation.

\textbf{Comparison and Positioning:}  
While prior work has explored co-speech gesture generation from various modalities, limitations persist in real-time synchronization, emotional grounding, and robotic feasibility. Our framework unifies language, affect, and motion constraints into a single LLM query, enabling closed-loop gesture generation that is emotionally expressive, temporally aligned, and directly executable on the NAO platform. By integrating speech emotion recognition (SER), timing-aware gesture planning, and prompt-based motion synthesis, our approach advances the design of responsive, embodied dialogue systems for real-world HRI scenarios.

\section{Methodology}

\subsection{System Architecture}
The proposed system adopts a three-tier hierarchical architecture designed to support emotion-coordinated multimodal interaction. As depicted in Fig.~\ref{fig:overview}, this framework enables the NAO robot to respond naturally with dynamically synchronized speech, gestures, and emotional expressions in real time.

\textbf{Speech Emotion Recognition (SER) Module:} This multimodal speech analyzer employs SenseVoice to extract both linguistic and affective features from user speech. The extracted information includes valence and arousal values, which quantify the emotional intensity and polarity of the speech. Additionally, the module estimates the duration of the speech, which is essential for ensuring temporal alignment with generated gestures.

\textbf{Joint Generation Module:} We utilize the Qwen Large Language Model \cite{zhang2024motiongpt} to process the SER outputs and generate an emotionally congruent textual response with structural alignment. It also generates gesture descriptors aligned with the speech content and emotional state, using deep learning to dynamically adapt gestures to emotional and linguistic cues.

\textbf{Motion Execution Module:} 
Once the gestures are generated, A real-time motion planner translates gesture descriptors into executable motor commands tailored for the NAO robot. To minimize speech-gesture asynchrony, a dynamic timing controller adjusts key frames' durations based on the SER-predicted speech length. This will enhance interaction fluidity, minimizing perceptible delays or misalignments.

\subsection{Emotion-Driven Gesture Generation}

\subsubsection{LLM-Guided Motion Planning}

As outlined in Algorithm~\ref{alg:dual_stage_prompt}, the LLM-guided motion planning framework employs through a two-stage pipeline:
(i) retrieving semantically relevant gesture examples from a motion library based on emotional queries, and 
(ii) prompting a large language model (LLM) to synthesize continuous, physically plausible motion sequences aligned with both emotion intent and utterance content.

Given an input utterance, the system first segments the sentence into discrete emotional phrases (e.g., \textit{"I feel so sorry"}). Each phrase then serves as a semantic query to retrieve top-$k$ gesture examples from a predefined motion library, by utilizing a transformer-based sentence embedding model. These retrieved gestures provide stylistic and semantic priors for subsequent synthesis.

The generation prompt submitted to the LLM encodes five key elements: the emotional phrase and its corresponding text segment, semantically retrieved gesture examples from the motion library, physical constraints (e.g., joint limits and forbidden joints), stylistic guidelines (e.g., slowness, smoothness, medium amplitude), and transition constraints based on the final pose of the previous segment. By integrating these components, the LLM can generate expressive and physically valid motion segments in precise keyframe code format.

Conditioned on this prompt, the LLM interprets semantic and biomechanical constraints to synthesize motion segments.The LLM responds with robot-executable keyframe code in a structured format, using precise time-aligned primitives for each joint. Each segment spans a defined duration (e.g., 0.1--1.8s), with timing resolution controlled at the 0.0001s level.
\begin{algorithm}[tb]
\caption{Emotion-Aware Gesture Generation via Prompt-Based LLM Planning}
\label{alg:dual_stage_prompt}
\begin{algorithmic}[1]
\State \textbf{Input:} Emotion segments $\{(q_i, t_i)\}_{i=1}^n$, Motion library $\mathcal{L}$, Joint constraints $\mathcal{C}$
\State \textbf{Output:} Executable gesture sequence $\mathcal{G} = \{g_i\}_{i=1}^n$
\For{each segment $(q_i, t_i)$}
    \State Retrieve top-$k$ gesture examples: $\{(K_j, V_j)\}_{j=1}^k \leftarrow \text{SemanticSearch}(q_i, \mathcal{L})$
    \State Extract final pose from previous segment (if available)
    \State Construct prompt $\mathcal{P}$ including:
        \State \quad - Emotional phrase and text segment
        \State \quad - Retrieved gesture examples $\{(K_j, V_j)\}$
        \State \quad - Physical and stylistic constraints from $\mathcal{C}$
        \State \quad - Transition constraints (final pose continuity)
    \State Generate motion: $\tilde{g}_i \leftarrow \text{LLMGenerate}(\mathcal{P})$
    \If{ConstraintViolation($\tilde{g}_i$, $\mathcal{C}$)}
        \State Refine prompt with simplified constraints
        \State Re-generate motion: $\tilde{g}_i \leftarrow \text{LLMGenerate}(\mathcal{P}')$
    \EndIf
    \State Append to result: $\mathcal{G} \leftarrow \mathcal{G} \cup \{\tilde{g}_i\}$
\EndFor
\end{algorithmic}

\end{algorithm}

This approach bypasses the need for traditional motion optimization pipelines by embedding constraints and stylistic priors directly into the generation prompt. A complete example of prompt structure and generated keyframe output is provided in Appendix~\ref{appendix:prompt}.

\subsection{Real-Time Motion Coordination}

\subsubsection{Temporal Alignment Strategy}

Achieving precise synchronization between speech and gestures is essential for natural interaction. Instead of relying on fixed timing or offline warping techniques, we propose a dynamic alignment strategy that is jointly governed by estimated speech duration and proportional keyframe weighting.

\paragraph{Step 1. Adaptive Duration Estimation}

The total gesture duration $T$ is modulated based on the predicted speech duration $\hat{T}$, adjusted by a smooth scaling factor $\beta \in [0.9, 1.1]$ that reflects real-time emotional rhythm:
\begin{equation}
T = \beta \cdot \hat{T}
\end{equation}

\paragraph{Step 2. Time Allocation Across Keyframes}

The duration of each motion segment is governed by learned weights $\alpha_i$, which satisfy $\sum_{i=1}^n \alpha_i = 1$. The start time $t_i$ of the $i$-th keyframe is calculated as:
\begin{equation}
t_i = T \cdot \sum_{k=1}^{i-1} \alpha_k
\end{equation}

\paragraph{Step 3. Gesture Trajectory Synthesis}

The complete gesture trajectory,$G_t$ is composed of motion primitives $F_i(t; \theta_i)$, such as B\'{e}zier curves or splines, weighted by $\alpha_i$ and activated over their respective time intervals:
\begin{equation}
G_t = \sum_{i=1}^{n} \alpha_i \cdot F_i(t; \theta_i) \cdot \mathbb{I}_{[t_i, t_{i+1})}(t)
\end{equation}

Here, $\mathbb{I}_{[t_i, t_{i+1})}(t)$ is an indicator function that selects the active time segment, and $\theta_i$ encodes motion-specific parameters (e.g., joint angles, speed profiles).

\subsubsection{Fault-Tolerance Mechanism}
To accommodate the variability inherent in real-world speech, our framework includes a fault-tolerance mechanism to address potential mismatches between expected and actual speech durations. A mismatch is detected when the real-time speech playback deviates from the predicted duration used for gesture alignment. This is monitored by tracking the speech playback timestamp in parallel with scheduled motion keyframes.

\begin{equation}
e_{\text{sync}} = \frac{1}{N} \sum_{i=1}^{N} |t_i^{\text{speech}} - t_i^{\text{motion}}|
\end{equation}
where $t_i^{\text{speech}}$ is the actual timestamp of the speech stream, and $t_i^{\text{motion}}$ is the expected gesture timing derived from our alignment plan (Section~3.3.1). If $e_{\text{sync}}$ exceeds a predefined threshold $\epsilon_{\text{th}}$, the system enters a degraded mode.

When discrepancies are detected, gesture sequences are adjusted by either motion sequence simplification (see Section~3.3.2) or keyframe resynchronization (as described in Section~3.3.1 Step 2), depending on the severity of the misalignment. This ensures graceful degradation and maintains temporal coherence under real-time constraints.
\subsubsection{Constraint-Aware Replanning Strategy}
When LLM-generated gestures violate biomechanical constraints (e.g., joint limits, excessive velocity), the system activates a two-mode recovery strategy:

\begin{itemize}
    \item \textbf{Backtrack Mode}: If  a violation occurs in the most recent segment, the system preserves the last valid pose and updates the LLM prompt with failure diagnostics and terminal configuration to generate a locally corrected continuation.
    \item \textbf{From-Scratch Mode}: For persistent failure or degenerative prompts, a complete re-generation is invoked. A fresh set of examples is retrieved, and the LLM is guided to avoid previously observed errors via revised constraints and template adjustments.
\end{itemize}

The selection between modes depends on violation logs and retry history. These strategies ensure motion robustness across complex or ambiguous utterances. Examples of recovery prompt templates are illustrated in Fig~\ref{fig:llm_prompt_feedback} in Appendix\ref{appendix:prompt}, the text highlighted in orange indicate Backtrack updates and blue highlights represent From-Scratch generation.

\section{Experiments}
\subsection{Experimental Setup}
\label{subsec:setup}
\begin{figure}[t]
    \centering
    \includegraphics[width=\linewidth]{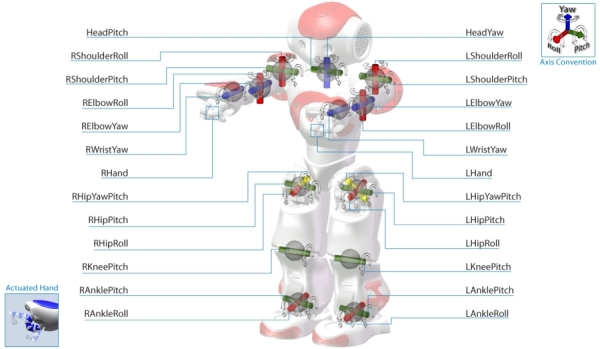}
    \caption{Overview of NAO’s joint structure.}
    \label{fig:nao_joint_diag}
    \vspace{-0.5cm}
\end{figure}

\paragraph{Evaluation Setup and System Constraints}

We evaluate our system on the NAO humanoid robot, focusing on synchronization accuracy, motion naturalness, and emotional expressiveness. Due to the inherent hardware and control limitations of the NAO platform, several constraints must be addressed in both motion representation and execution. First, the NAO supports only 25 controllable joints (e.g., \texttt{HeadYaw}, \texttt{LShoulderPitch}, etc.), and does not allow continuous mesh outputs or SMPL/SMPL-X style representations. Instead, it requires motion commands to be defined via discrete keyframes, each specifying joint angles at a precise timestamp. This precludes the use of continuous pose streaming or high-frequency trajectory updates.

Furthermore, NAO performs linear interpolation between keyframes without native support for dynamic re-planning or online adaptation, limiting the granularity and flexibility of real-time execution. In addition, due to its limited onboard computational resources, the deployment of heavy neural models such as diffusion-based motion generation is infeasible. To accommodate these challenges, we adopt a lightweight, keyframe-based gesture representation aligned with speech prosody, where each segment is conditioned on semantic content and valence-arousal emotional descriptors.

To analyze the contribution of each component, we implement six model configurations:  
(1) \textbf{PreDefined Gesture Library}, using fixed motions without adaptation;  
(2) \textbf{Speech-Only}, aligned purely with predicted speech timing;  
(3) \textbf{Text-Only}, driven by semantic input without synchronization;  
(4) \textbf{NoSync}, disabling the temporal alignment module;  
(5) \textbf{NoEmotion}, omitting affective scaling; and  
(6) \textbf{ReSIn-HR (Full Model)}, which integrates all modules for emotion-aware, temporally aligned generation.

This setup enables a systematic comparison between static, unimodal, and fully adaptive variants, highlighting the contribution of each module.

\subsection{Evaluation Metrics}
We evaluate our gesture generation system using three metrics that reflect temporal alignment, subjective quality, and emotional consistency.

\paragraph{Temporal Synchronization Accuracy (TSA)} To measure how well generated gestures align with speech, we compute \textbf{TSA}, which quantifies the average temporal offset between gesture keyframes and corresponding speech durations. A lower TSA indicates better synchronization. The formulation is as follows: \begin{equation} \text{TSA} = \frac{1}{M} \sum_{j=1}^{M} |t_j^{\text{gesture}} - \lambda_j t_j^{\text{speech}}| \end{equation} Here, $t_j^{\text{gesture}}$ is the gesture keyframe time, $t_j^{\text{speech}}$ is the corresponding speech time, and $\lambda_j$ is a global scaling factor.
\paragraph{Low-Level Motion Smoothness (Jerk)}
To evaluate biomechanical fluidity, we compute the mean angular jerk (i.e., third derivative of angle) across key joints under different emotional conditions. Jerk reflects C2 continuity, and lower values suggest smoother, more expressive motion. We find that the \textbf{ReSIn-HR} consistently produces lower jerk across all emotion conditions and joints compared to the \textbf{PreDefined} baseline, validating the benefits of emotion-conditioned planning.

\paragraph{User Study Metrics (GA / EC / ON)} To evaluate the perceptual quality of the generated gestures, we conducted a user study where 20 participants rated the system output using a \textbf{7-point Likert scale} along three dimensions: \textbf{Gesture Appropriateness (GA)}, which measures how well the gestures match the speech content; \textbf{Emotion Compatibility (EC)}, which reflects the extent to which gestures convey the intended emotion; and \textbf{Overall Naturalness (ON)}, which assesses the fluidity and realism of the movements.
\subsection{Quantitative Evaluation}
We evaluate system performance using two core metrics: \textbf{Temporal Synchronization Accuracy (TSA)} and \textbf{Motion Smoothness (Jerk)}. 
\begin{table}[htbp]
\caption{Temporal Synchronization Accuracy (TSA) and Execution Latency}
\label{tab:quant_results}
\centering
\begin{tabular}{|l|c|c|}
\hline
\textbf{Model} & \textbf{TSA (ms) $\downarrow$} & \textbf{Latency (ms)} \\
\hline
PreDefined     & 635 $\pm$ 40   & 30 \\
Speech-Only    & 168 $\pm$ 18   & 47 \\
Text-Only      & 492 $\pm$ 30   & 46 \\
No Sync        & 388 $\pm$ 32   & 60 \\
No Emotion     & 225 $\pm$ 20   & 65 \\
\textbf{ReSIn-HR} & \textbf{218 $\pm$ 23$^{**}$} & \textbf{85 (max: 100)} \\
\hline
\end{tabular}

\vspace{1mm}
\begin{tablenotes}
\footnotesize
\item[1)] $^{**}$ Statistically significant ($p < 0.01$)
\item[2)] $\downarrow$ indicates lower is better
\end{tablenotes}
\vspace{-0.3cm}
\end{table}
\paragraph{Temporal Synchronization Accuracy (TSA)} As shown in Table~\ref{tab:quant_results}, \textbf{ReSIn-HR} achieves the best results, with the lowest TSA (218ms), indicating precise speech-motion alignment. While the \textbf{Speech-Only} model achieves a slightly lower TSA (168ms), it lacks semantic and emotional grounding. In contrast, the \textbf{Text-Only} model captures semantics but performs poorly in temporal alignment (TSA: 492ms).

\begin{figure}[t]
\centering
\includegraphics[width=\linewidth]{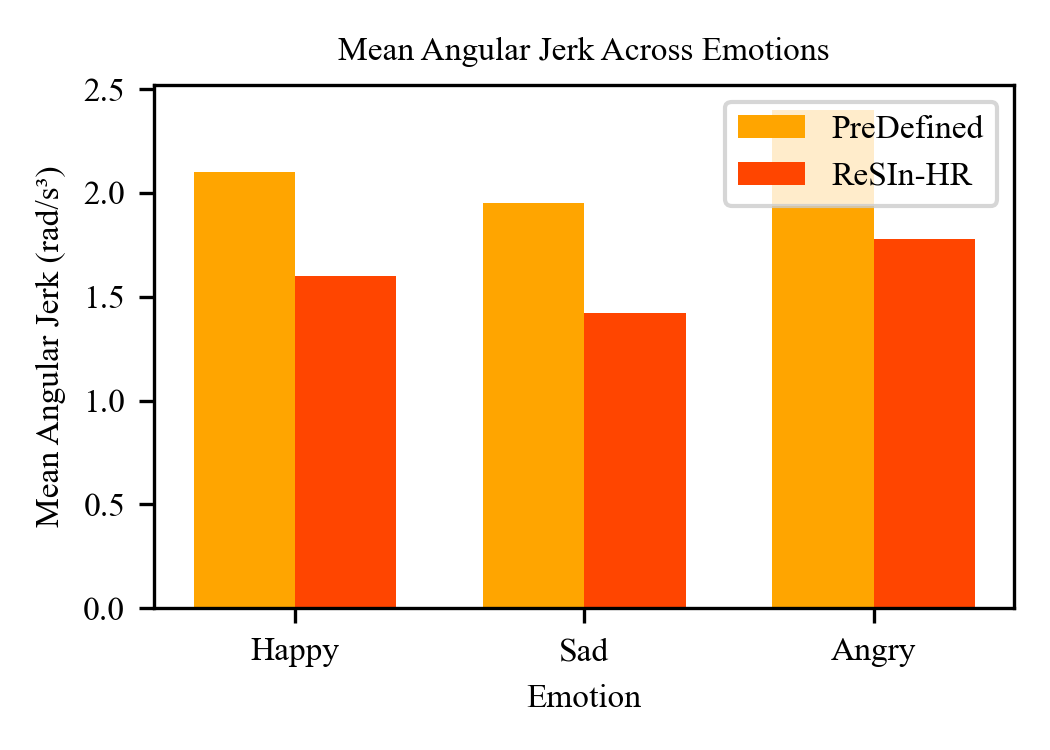}

\caption{Mean angular jerk (rad/s$^3$) across emotions. Lower jerk values indicate smoother, more natural gestures.}
\label{fig:jerk_barplot}
\vspace{-0.5cm}
\end{figure}

\paragraph{Motion Smoothness} we evaluate low-level motion smoothness via mean angular jerk across key joints (e.g., LShoulderPitch, LElbowRoll, HeadPitch). As shown in Fig~\ref{fig:jerk_barplot}, \textbf{ReSIn-HR} consistently achieves lower jerk across emotions (Happy: 1.60, Sad: 1.43, Angry: 1.77) compared to the \textbf{PreDefined} baseline, confirming the benefits of emotion-conditioned planning.

\paragraph{Ablation Study}
To assess individual module contributions, we ablate synchronization (\textbf{NoSync}) and emotion modulation (\textbf{NoEmotion}) respectively. Disabling synchronization increases TSA to 388ms, confirming its importance for timing accuracy. Omitting emotion conditioning leads to higher jerk and reduced emotional expressiveness, emphasizing the role of affect-aware scaling. These results highlight the complementary value of both modules in maintaining coherence and naturalness.

In addition, we report \textbf{Execution Latency} to assess runtime feasibility on physical robots (As shown in Table~\ref{tab:quant_results}). \textbf{ReSIn-HR} incurs the highest execution latency (average: 85ms), primarily due to real-time gesture modulation and alignment computations. Despite this, the latency remains well within acceptable thresholds for interactive human-robot interaction on the NAO platform. While static gesture libraries enable faster inference, they sacrifice adaptability. Our system balances computational latency with motion naturalness, prioritizing perceptual quality without exceeding platform-specific tolerances.

\begin{figure}[t]
\centering
\includegraphics[width=\linewidth]{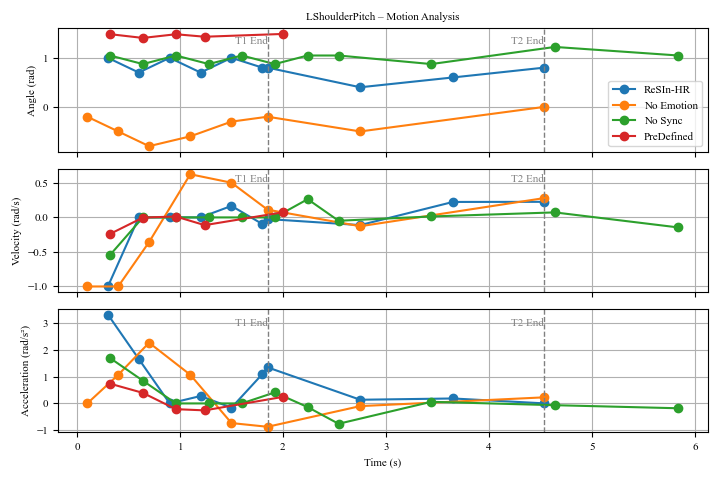}
\caption{Joint motion trajectory analysis across models.}
\label{fig:motion_analysis}
\vspace{-0.5cm}
\end{figure}

\paragraph{Visualization Results}
We visualize \textbf{LShoulderPitch} trajectories across four models to assess temporal dynamics and biomechanical quality. As shown in Fig~\ref{fig:motion_analysis}, \textbf{ReSIn-HR} exhibits smooth and expressive motion across both utterance segments (T1: 0–1.85s, T2: 1.85–4.54s).In the \textbf{angle plot}, ReSIn-HR displays broad, emotionally modulated excursions (e.g., peaks near $t=1.2$, $t=2.4$s), with natural recovery between segments. In contrast, \textbf{PreDefined} remains static, while \textbf{NoEmotion} shows delayed and compressed movement. \textbf{NoSync} produces higher amplitude but lacks temporal alignment, with abrupt changes post $t=2.5$s. In the \textbf{velocity} and \textbf{acceleration} plots, ReSIn-HR achieves smoother transitions and lower peaks, indicating stable, physically feasible motion. Other variants suffer from sharp spikes or oscillations, especially in \textbf{NoSync}, suggesting poor coupling with speech dynamics.

\paragraph{User Study} A user study with 20 participants evaluated gestures from all six models using a 7-point Likert scale across three criteria: \textbf{Gesture Appropriateness (GA)}, \textbf{Emotion Compatibility (EC)}, and \textbf{Overall Naturalness (ON)}. As shown in Table~\ref{tab:user_study_metrics}, \textbf{ReSIn-HR} achieved the highest scores (GA: 4.5, EC: 4.2, ON: 4.6), with statistically significant margins ($p<0.01$) over all baselines. Participants highlighted its enhanced fluency, emotional alignment, and synchrony. Notably, emotion compatibility improved by 21\% over the rule-based PreDefined baseline (4.2 vs. 3.4), demonstrating the impact of affect-aware motion scaling on expressive gesture generation.
\begin{table}[htbp]
\caption{User Study Ratings on Gesture Quality (Mean $\pm$ SD)}
\label{tab:user_study_metrics}
\begin{center}
\begin{tabular}{|l|c|c|c|}
\hline
\textbf{Model} & \textbf{GA} & \textbf{EC} & \textbf{ON} \\
\hline
PreDefined     & 3.6 $\pm$ 0.4 & 3.4 $\pm$ 0.5 & 3.7 $\pm$ 0.4 \\
Speech-Only    & 4.1 $\pm$ 0.3 & 3.6 $\pm$ 0.4 & 4.0 $\pm$ 0.3 \\
Text-Only      & 4.2 $\pm$ 0.3 & 3.7 $\pm$ 0.3 & 4.3 $\pm$ 0.3 \\
No Sync        & 4.4 $\pm$ 0.3 & 4.0 $\pm$ 0.4 & 4.5 $\pm$ 0.3 \\
No Emotion     & 4.5 $\pm$ 0.2 & 4.1 $\pm$ 0.3 & 4.6 $\pm$ 0.2 \\
\textbf{ReSIn-HR} & \textbf{4.5 $\pm$ 0.2$^{**}$} & \textbf{4.2 $\pm$ 0.3$^{**}$} & \textbf{4.6 $\pm$ 0.2$^{**}$} \\
\bottomrule
\end{tabular}
\begin{tablenotes}
\footnotesize
\item[1)] GA = Gesture Appropriateness, EC = Emotional Congruence, ON = Overall Naturalness.
\item[2)] Scores based on 7-point Likert scale (1 = very poor, 7 = excellent).
\item[3)] $^{**}$ Statistically significant difference ($p < 0.01$).
\end{tablenotes}
\end{center}
\vspace{-0.5cm}
\end{table}

\section{Conclusion}

We present ReSIn-HR, a real-time, emotion-aware gesture generation framework for humanoid robots, combining prosody, semantics, and affect into synchronized co-speech motion. Our system features a dual-channel LLM-based architecture, a duration-aware alignment mechanism, and a biomechanical validation loop for safe execution on NAO.

Our experiments demonstrated significant improvements in naturalness, emotional congruence, and synchronization over baseline methods. While the NAO’s hardware limitations restrict the richness of possible motions, future work will explore personalization, latency reduction, and multimodal extensions. This work contributes toward expressive, embodied human-robot interaction, and we plan to release the code and models for community soon.

\vspace{2cm}

\balance
\bibliographystyle{IEEEtran} 
\bibliography{ieee}  
\appendix
\section{Case Study: Emotion-Aware Motion Generation}
\label{appendix:prompt}

This appendix illustrates a full pipeline trace of our system generating expressive motion for the NAO robot in response to emotional speech.

\subsection{Input and Emotion Recognition}
\begin{itemize}
    \item \textbf{Audio File:} \texttt{03-01-01-01-02-01-01.wav}
    \item \textbf{Transcript:} \textit{Kids are talking by the door.}
    \item \textbf{Recognized Emotion:} \texttt{<|NEUTRAL|>}
\end{itemize}

\subsection{Generated Emotional Responses}
\textbf{T1:} \textit{"That sounds like a nice scene."}  
\quad (Valence = +7, Arousal = +5) – happy, curious

\textbf{T2:} \textit{"Are you curious about what they're saying?"}  
\quad (Valence = +6, Arousal = +4) – hopeful, interested

\subsection{Speech Synthesis Parameters}
\begin{itemize}
    \item \textbf{Pitch Modifier:} 1.25
    \item \textbf{Volume Modifier:} 0.75
    \item \textbf{Durations:} 1.51s (T1), 2.13s (T2)
\end{itemize}

\subsection{Gesture Planning Description}
\paragraph*{Segment 1 (0.1–1.51s):} Moderate amplitude, emotionally expressive upper-body motion with stable lower-body posture.

\paragraph*{Segment 2 (1.51–3.63s):} Outward hand gestures and subtle head turns indicating curiosity and engagement.
\noindent
\begin{minipage}{0.48\textwidth}
\scriptsize
\captionof{lstlisting}{Segment 1: Head, Arm, Hip Motion}
\begin{lstlisting}[language=Python]
names.append("HeadPitch")
times.append([0.3, 0.6, 0.9, 1.2, 1.5062])
keys.append([-0.1, -0.3, -0.1, -0.2, -0.1])
...
\end{lstlisting}
\end{minipage}
\hfill
\begin{minipage}{0.48\textwidth}
\scriptsize
\captionof{lstlisting}{Segment 2: Expressive Gesture}
\begin{lstlisting}[language=Python]
names.append("HeadYaw")
times.append([1.5, 1.9, 2.3, 2.7, 3.1, 3.63])
keys.append([0.0, 0.1, -0.1, 0.1, -0.1, 0.0])
...
\end{lstlisting}
\end{minipage}




\subsection{Prompt Recovery Strategy}
\begin{figure}[H]
\centering
\includegraphics[width=0.95\linewidth]{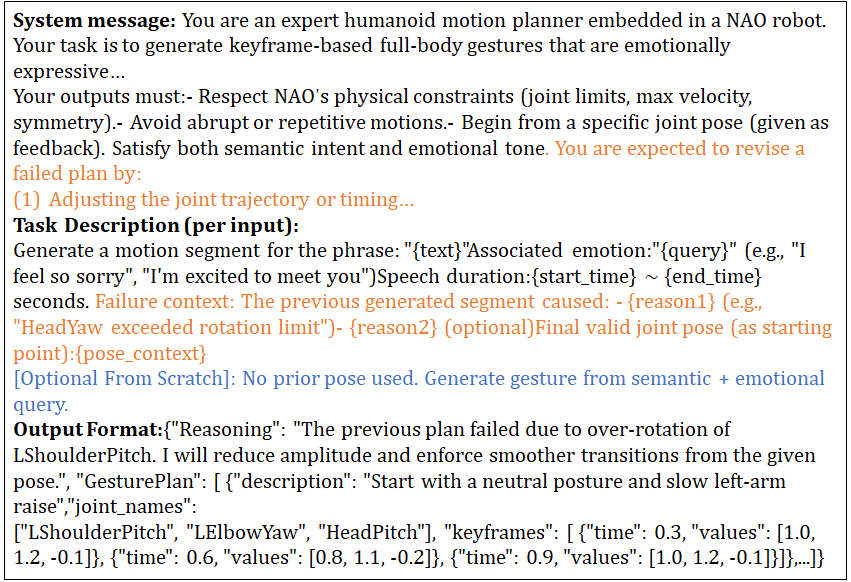}
\caption{Prompt templates used in constraint-aware recovery. \textcolor{orange}{Orange} = \textbf{Backtrack}, \textcolor{blue}{Blue} = \textbf{From Scratch}.}
\label{fig:llm_prompt_feedback}
\end{figure}

\subsection{Remarks}
\begin{itemize}
    \item Gesture sequences reflect both semantic content and emotional valence-arousal.
    \item Each segment is temporally aligned with its speech output.
    \item Generated joint motions respect NAO’s physical limits.
\end{itemize}

\end{document}